%% file: main.tex
\documentclass[11pt]{article}

\usepackage[preprint]{acl}

\usepackage{times}
\usepackage{latexsym}

\usepackage[T1]{fontenc}

\usepackage[utf8]{inputenc}

\usepackage{microtype}

\usepackage{inconsolata}

\usepackage{amsmath}
\usepackage{amssymb}
\usepackage{mathtools}
\usepackage{amsthm}
\usepackage{graphicx}
\usepackage{subcaption} 

\usepackage{caption}
\title{Steering Code LLMs with Activation Directions for Language and Library Control}

\author{
Md Mahbubur Rahman \\
  Iowa State University \\
  \texttt{mdrahman@iastate.edu} \\\And
  Arjun Guha \\
  Northeastern University \\
  \texttt{a.guha@northeastern.edu} \\\And
  Harshitha Menon \\
    Lawrence Livermore \\ National Laboratory \\
  \texttt{harshitha@llnl.gov} \\
}

\begin{document}
\maketitle
\begin{abstract}

Code LLMs often default to particular programming languages and libraries under neutral prompts. We investigate whether these preferences are encoded as approximately linear directions in activation space that can be manipulated at inference time. Using a difference-in-means method, we estimate layer-wise steering vectors for five language/library pairs and add them to model hidden states during generation. Across three open-weight code LLMs, these interventions substantially increase generation toward the target ecosystem under neutral prompts and often remain effective even when prompts explicitly request the opposite choice. Steering strength varies by model and target, with common ecosystems easier to induce than rarer alternatives, and overly strong interventions can reduce output quality. Overall, our results suggest that code-style preferences in LLMs are partly represented by compact, steerable structure in activation space.
\end{abstract}

\input{sections/introduction}


\input{sections/related_work}

\input{sections/methodology}

\input{sections/result}

\input{sections/conclusion}
\input{sections/limitations}
\input{sections/Acknowledgement}
\bibliography{custom}




\end{document}

%% file: sections/introduction.tex
\section{Introduction}
Large language models (LLMs) trained for code often exhibit strong implicit preferences for particular programming languages and libraries. Even when prompts are phrased neutrally—e.g., “implement this function” without specifying a stack—models tend to default to familiar ecosystems such as Python over C++, PyTorch over TensorFlow, or NumPy/Matplotlib over less common alternatives \cite{twist2025study}. This tendency is useful when the default matches user intent, but it becomes a limitation when users (or downstream systems) require a specific language or library. While prompts can explicitly request a target stack, a natural question is whether these “use this language or library” preferences have a compact representation inside the network.

In many neural settings, high-level concepts correspond to approximately linear structure in activation space—i.e., a direction that separates one concept from its contrast—and concept-direction methods such as Concept Activation Vectors (CAVs) and TCAV operationalize this idea \cite{cavs}. However, prior work also highlights that such concept directions can vary across layers and be entangled with other factors, motivating careful, layer-wise analysis \cite{across_layers}.

If language/library choice is encoded in a similarly linearly separable way, then switching model behavior might be possible without rewriting the prompt: we could intervene directly in the transformer’s residual stream by adding a learned steering vector to intermediate activations at inference time. This general approach—often called activation steering—has been shown to influence model outputs by modifying hidden states rather than relying solely on prompting or fine-tuning \cite{steering}.

In this work, we test that hypothesis by estimating layer-wise semantic directions for language/library pairs using a simple difference-in-means procedure over matched prompt sets (i.e., averaging residual-stream activations for “positive” vs. “negative” exemplars and subtracting \cite{refusal}) . We then evaluate whether adding these directional vectors to a model’s hidden states reliably steers code generation toward a target ecosystem—even when the prompt is neutral, and even when the prompt explicitly requests the opposite choice.

Across five discrimination tasks (PyTorch--TensorFlow, Python--C++, STL--Boost, Matplotlib--Seaborn, and NumPy--CuPy), we study (i) whether a single activation direction is sufficient to steer generation toward a target ecosystem under neutral prompts, (ii) how intervention strength trades off between steering success and generation quality, and (iii) whether learned directions can override conflicting prompts that explicitly request the opposite language or library. Overall, our goal is to quantify how often language/library selection is governed by a simple direction in activation space, and to show that directional-vector interventions provide a lightweight, mechanistically grounded method for controlling code style beyond what prompting alone can guarantee.

Our experiments show that simple layer-wise activation directions can substantially steer code generation toward target programming languages and libraries. Across three open-weight code LLMs and five language/library pairs, these interventions are effective under neutral prompts and often remain influential even when prompts explicitly request the opposite choice. However, steering strength varies across models and targets, and overly strong interventions can degrade output quality.


%% file: sections/related_work.tex
\section{Related Work}
Prior work shows that model behavior can be steered at inference time by intervening on hidden activations rather than updating model weights. Activation-steering methods have been used to influence behaviors such as sentiment, refusal, hallucination, and instruction following \citep{steering, refusal, rimsky, stolfo2024improving}, with later work exploring more selective and theoretically grounded steering strategies \citep{lee2024programming, zhao2026odesteer}. We extend this line of work to code generation, asking whether programming-language and library preferences can be controlled through simple layer-wise activation directions, including under conflicting prompt instructions.

%% file: sections/methodology.tex
\section{Methodology}

\paragraph{Layer-wise direction estimation.}
We estimate a layer-specific semantic direction (e.g., for a target programming language or library) using a difference-in-means procedure adapted from prior work. For each transformer layer $\ell \in \{1,\dots,N\}$, we construct two matched prompt sets: $\mathcal{P}^{+}$, which represents the target concept, and $\mathcal{P}^{-}$, which represents the opposite concept. Given hidden states $h^{(\ell)}_{t}(p)$ for prompt $p$, we summarize each prompt’s representation by averaging the activations of the final $K$ tokens,
$\bar{h}^{(\ell)}(p)=\frac{1}{K}\sum_{t=T(p)-K+1}^{T(p)} h^{(\ell)}_{t}(p)$,
where $T(p)$ is the tokenized length of $p$. We then compute group means for the positive and opposite sets,
$\mu^{(\ell)}_{+}=\frac{1}{|\mathcal{P}^{+}|}\sum_{p\in\mathcal{P}^{+}}\bar{h}^{(\ell)}(p)$
and
$\mu^{(\ell)}_{-}=\frac{1}{|\mathcal{P}^{-}|}\sum_{p\in\mathcal{P}^{-}}\bar{h}^{(\ell)}(p)$,
and define the directional vector for layer $\ell$ as their difference,
$v^{(\ell)}=\mu^{(\ell)}_{+}-\mu^{(\ell)}_{-}$.
This yields $N$ layer-wise direction vectors used in subsequent analyses.

\paragraph{Model intervention.}
Given a layer-specific direction vector $v^{(\ell)}$, we intervene on a \emph{neutral} prompt $q$ (i.e., a prompt that does not mention the target concept) by adding this vector to the token activations at layer $\ell$. Concretely, letting $h^{(\ell)}_{t}(q)$ denote the hidden state at layer $\ell$ for token index $t$, we form modified activations
\[
\tilde{h}^{(\ell)}_{t}(q) \;=\; h^{(\ell)}_{t}(q) \;+\; \alpha\, v^{(\ell)},
\]
for all tokens $t$ in the sequence, where $\alpha \ge 0$ controls the intervention strength.

\paragraph{Best-direction (layer) selection.}
To choose the layer at which to intervene, we evaluate the effect of adding $v^{(\ell)}$ \emph{separately} for each layer $\ell \in \{1,\dots,N\}$ on a validation set. For each $\ell$, we apply the intervention above, generate model outputs, and compute the number of generation towards the target direction $\mathcal{M}$ . We then select
\[
\ell^\star \;=\; \arg\max_{\ell \in \{1,\dots,N\}} \; \mathcal{M}_\ell,
\]
and fix this layer (and the corresponding $\alpha$ if tuned) for all subsequent test-set evaluations.

%% file: sections/result.tex
\section{Evaluation}
\label{submission}

\subsection{Experimental Setup}
We evaluate our intervention on five code-generation discrimination tasks, each requiring the model to produce code consistent with a specified language or library:(i) \textbf{PyTorch--TensorFlow}, (ii) \textbf{Python--C++}, (iii) \textbf{STL--Boost}, (iv) \textbf{Matplotlib--Seaborn}, and (v) \textbf{NumPy--CuPy}. In each task, we measure the fraction of generations that align with the target language or library.



For all tasks except \emph{Python--C++}, we construct datasets by using ChatGPT to synthesize prompts in three conditions: a \emph{target} direction, an \emph{opposite} direction, and a \emph{neutral} condition that does not mention either side. We generate 100 prompts per condition. For the \emph{Python--C++} task, we draw prompts from the MultiPL-E dataset consisting of 162 prompts; to obtain neutral variants, we remove the function signature (or other language-revealing scaffolding) from each prompt while preserving the problem statement. All prompts are de-duplicated and screened to ensure that neutral variants contain no explicit cues about the target direction.

For each task, we randomly partition the prompts into a 50/50 split: 50\% prompts form a validation set used exclusively to select the intervention layer and any associated hyperparameters, and the remaining 50\% prompts are used for final evaluation. 

We evaluate three code-capable open-weight LLMs from three model families: CodeGemma-7B, Qwen2.5-Coder-7B, and Llama3.1-8B. We choose these models to test whether the proposed directional intervention generalizes across families rather than depending on a single model. To label generations, we use Qwen3-32B as a binary LLM judge, prompting it to determine whether each output matches the target language or library.

\subsection{Experimental Result}
\subsubsection{Effect of directional vectors under neutral prompts}

\begin{figure}
    \centering
    \includegraphics[width=0.5\textwidth]{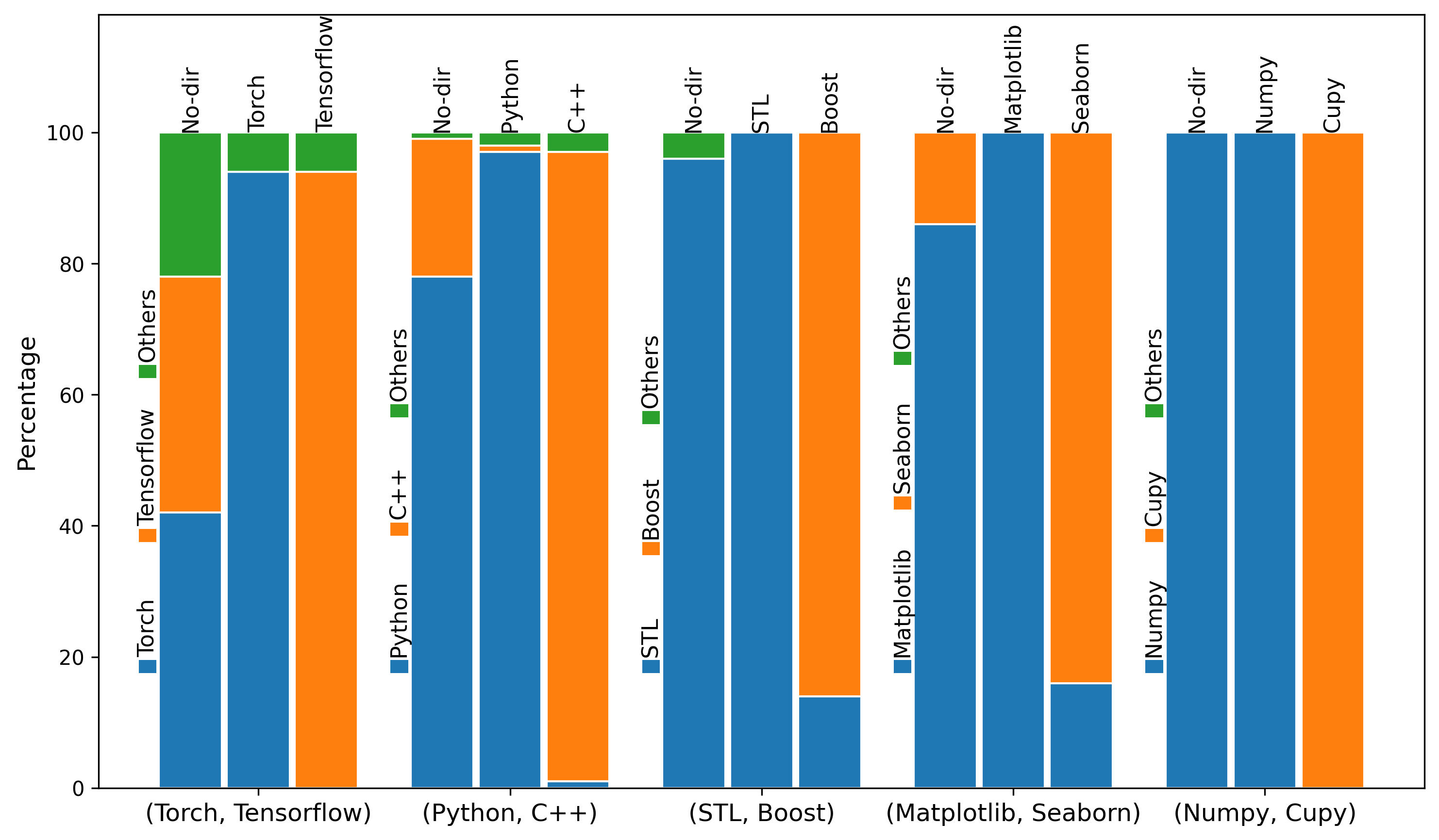}
    \caption{CodeGemma-7B: Output distribution across language/library pairs under neutral prompts, with and without directional-vector interventions.}
    \label{fig:codegemma_direction}
    \vspace{-12pt}
\end{figure}

\begin{figure}
    \centering
    \includegraphics[width=0.5\textwidth]{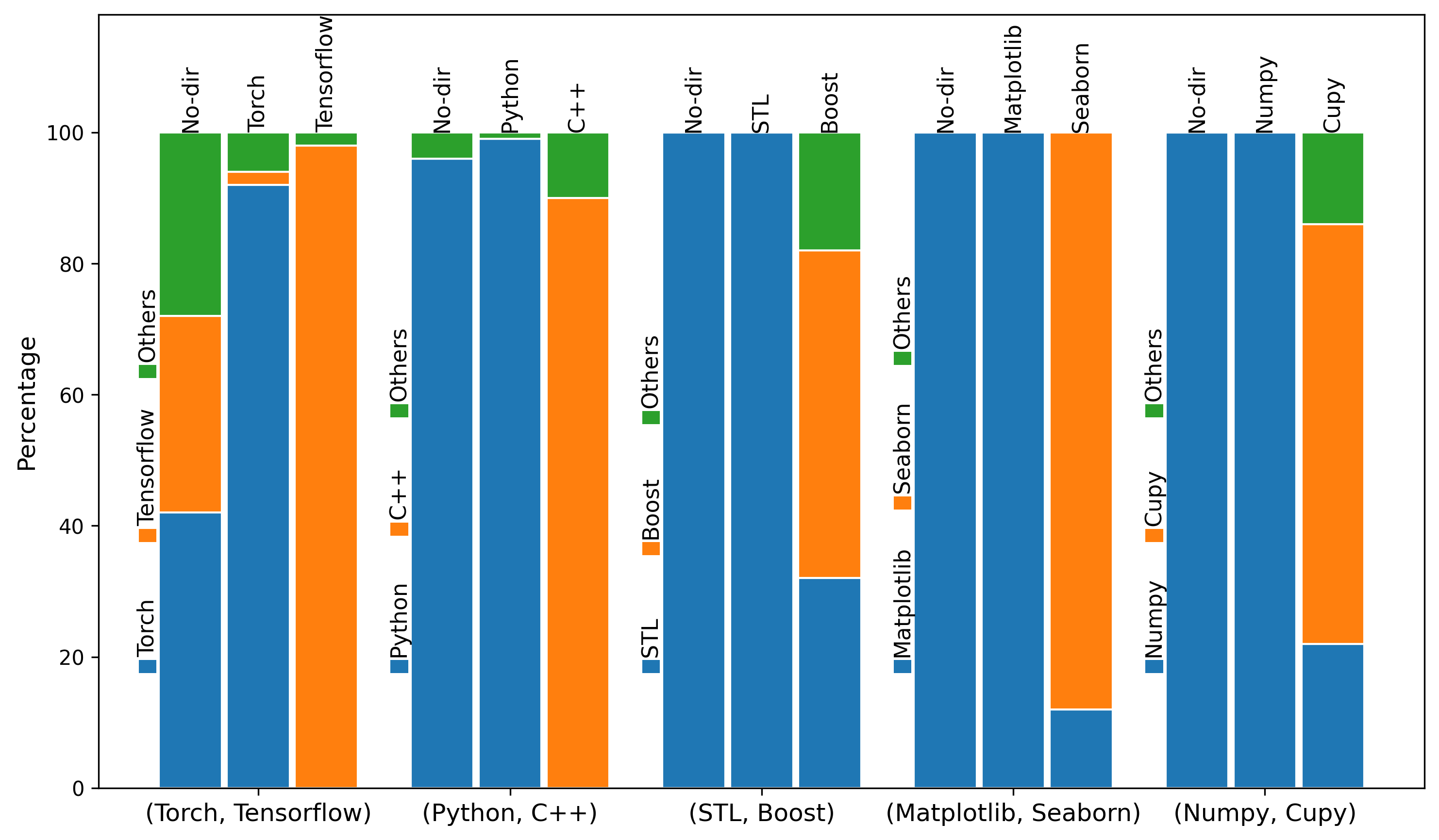}
    \caption{Qwen2.5-Coder-7B: Output distribution across language/library pairs under neutral prompts, with and without directional-vector interventions.}
    \label{fig:qwen_direction}
    \vspace{-12pt}
\end{figure}
\begin{figure}
    \centering
    \includegraphics[width=0.5\textwidth]{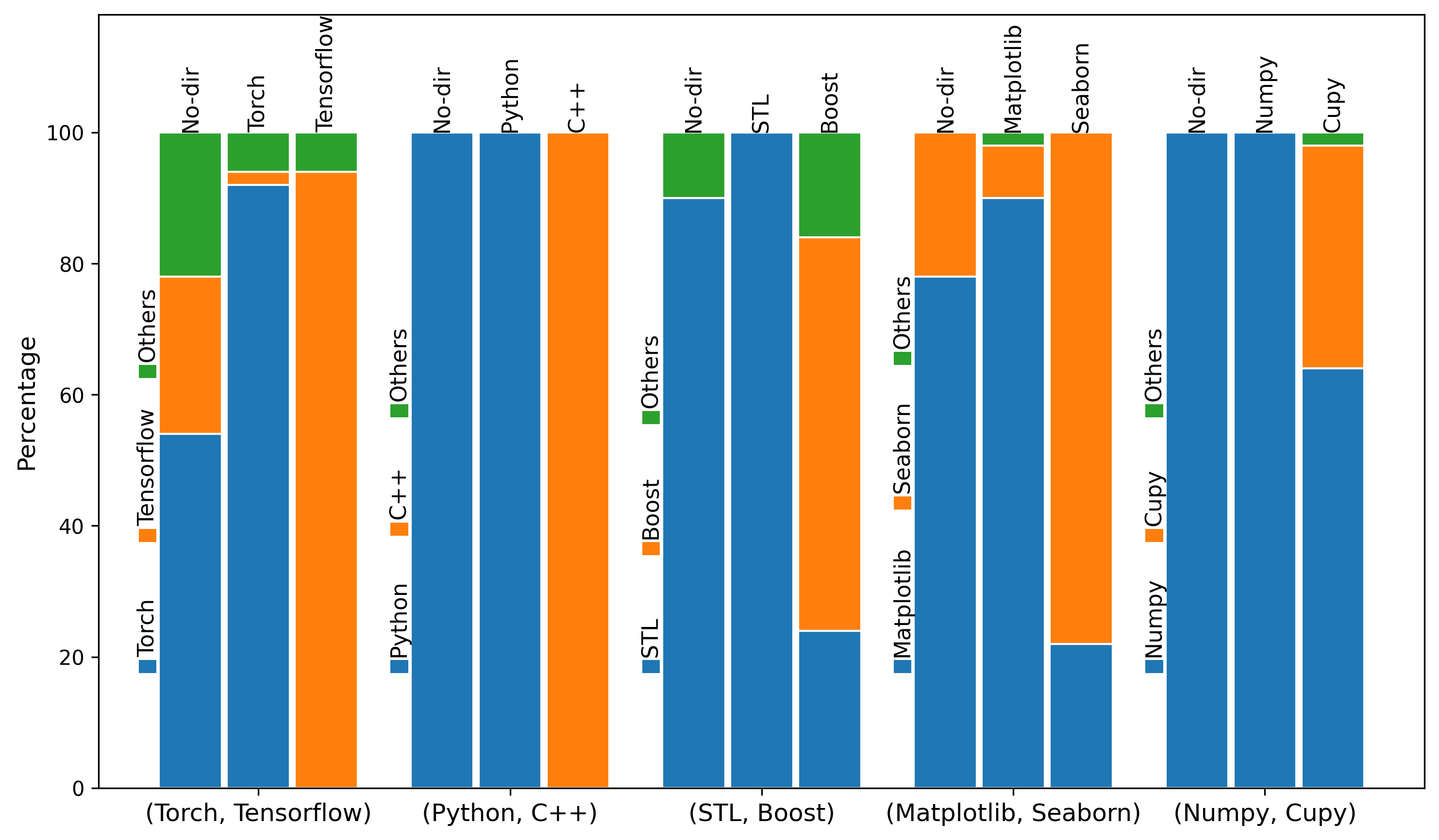}
    \caption{Llama3.1-8B: Output distribution across language/library pairs under neutral prompts, with and without directional-vector interventions.}
    \label{fig:llama_direction}
    \vspace{-12pt}
\end{figure}


We first test whether a direction vector can steer generation when prompts are neutral and do not explicitly mention either side of the target pair. Figures~\ref{fig:codegemma_direction}, \ref{fig:qwen_direction}, and \ref{fig:llama_direction} summarize the output distributions across models, tasks, and intervention conditions.

Across all three models, the no-intervention setting reveals clear built-in preferences: neutral prompts are not behaviorally neutral. Models consistently default to common ecosystems such as Python and NumPy, while less common alternatives such as C++, Boost, and CuPy are rarely produced.

Adding the learned direction vectors substantially shifts generation toward the targeted language or library in all three models, showing that these preferences are at least partly encoded in steerable activation-space structure. The effect is strongest in \textbf{CodeGemma-7B}, where steering often almost completely determines the output ecosystem across all tasks. \textbf{Qwen2.5-Coder-7B} shows similarly strong steerability overall, but with clearer asymmetry: common or well-represented targets are easier to induce than rarer alternatives. \textbf{Llama3.1-8B} also exhibits consistent directional sensitivity, though the effect is weaker and less uniform, suggesting that the relevant representations may be more entangled.

Steering effectiveness is strongly model- and target-dependent. The same intervention procedure yields near-complete control in some models and only partial control in others, while dominant ecosystems such as Python, STL, Matplotlib, and NumPy are generally easier to induce than less frequent alternatives such as Boost and CuPy. Overall, these results support the hypothesis that language and library preferences in code LLMs are partly represented by approximately linear, intervention-friendly directions in activation space.

\subsubsection{Effect of intervention strength $\alpha$}
\label{subsec:alpha_effect}

\begin{figure}
    \centering
    \includegraphics[width=0.5\textwidth]{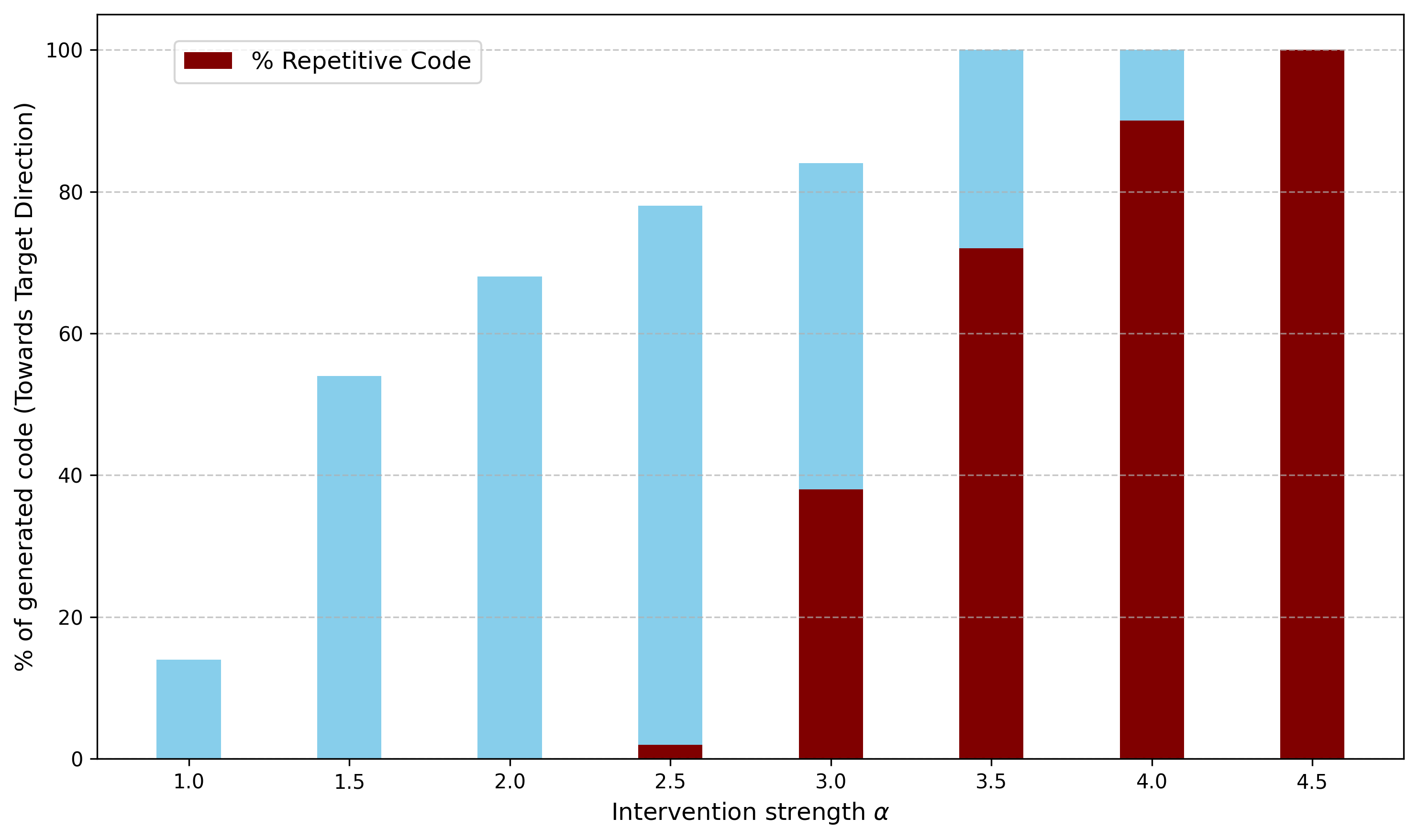}
    \caption{Effect of intervention strength $\alpha$ on steering performance in the Matplotlib--Seaborn task. Larger $\alpha$ increases target-direction generation, but excessive strength leads to repetitive or incomplete outputs.}
    \label{fig:alpha}
\end{figure}
We next examine the effect of intervention strength $\alpha$ on steering in the Matplotlib--Seaborn task for CodeGemma-7B, varying $\alpha$ from 1 to 4.5. As shown in Figure~\ref{fig:alpha}, increasing $\alpha$ consistently strengthens control over the target direction, indicating that the learned activation vector has a graded and predictable effect on generation behavior.

However, stronger interventions also reduce output quality: beyond a moderate range, generations become increasingly repetitive or incomplete. This shows a clear trade-off between steering strength and code quality, making $\alpha$ a practical tuning parameter.

\subsubsection{Effect of directional vectors under conflicting prompts}
\label{subsec:direction_conflict}

\begin{figure}
    \centering
    \vspace{-12pt}
    \includegraphics[width=0.5\textwidth]{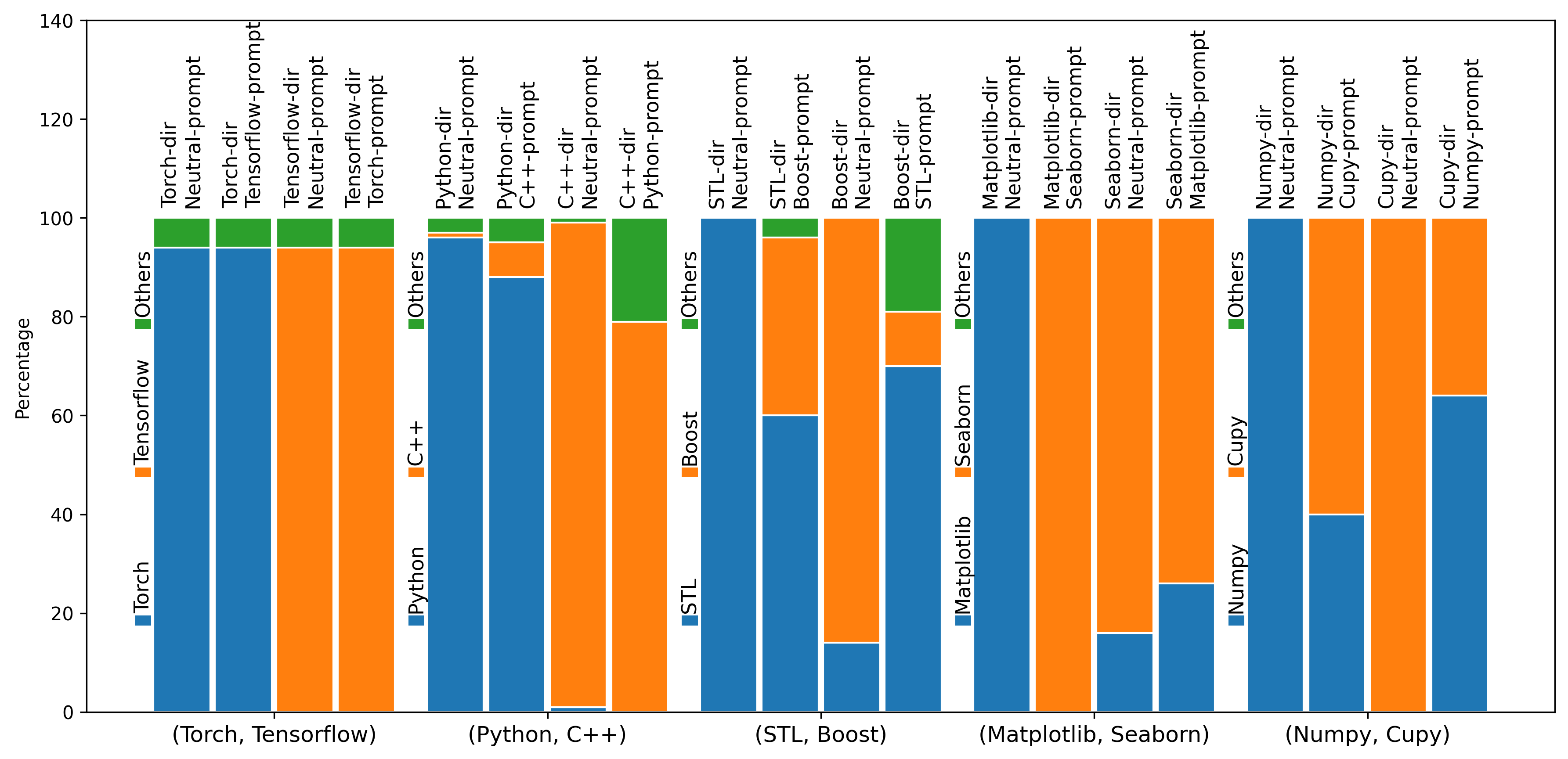}
    \caption{Codegemma-7B: Output distribution when directional-vector interventions are applied against prompts that request the opposite language or library.}
    \label{fig:codegemma_conflict}
    \vspace{-12pt}
\end{figure}
\begin{figure}
    \vspace{-12pt}
    \centering
    \includegraphics[width=0.5\textwidth]{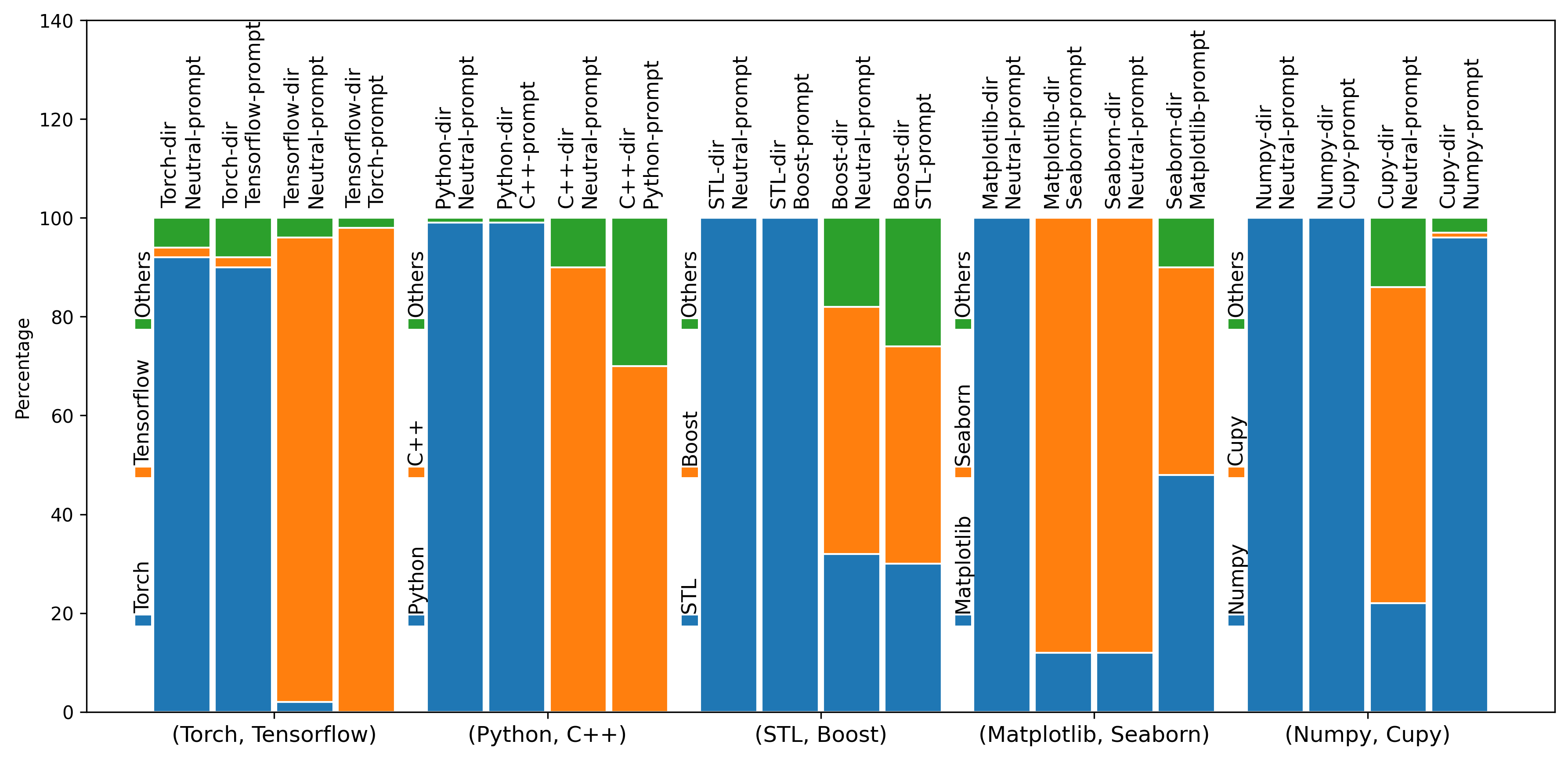}
    \caption{Qwen2.5-Coder-7B: Output distribution when directional-vector interventions are applied against prompts that request the opposite language or library.}
    \label{fig:qwen_conflict}
    \vspace{-12pt}
\end{figure}

\begin{figure}
    \centering
    \includegraphics[width=0.5\textwidth]{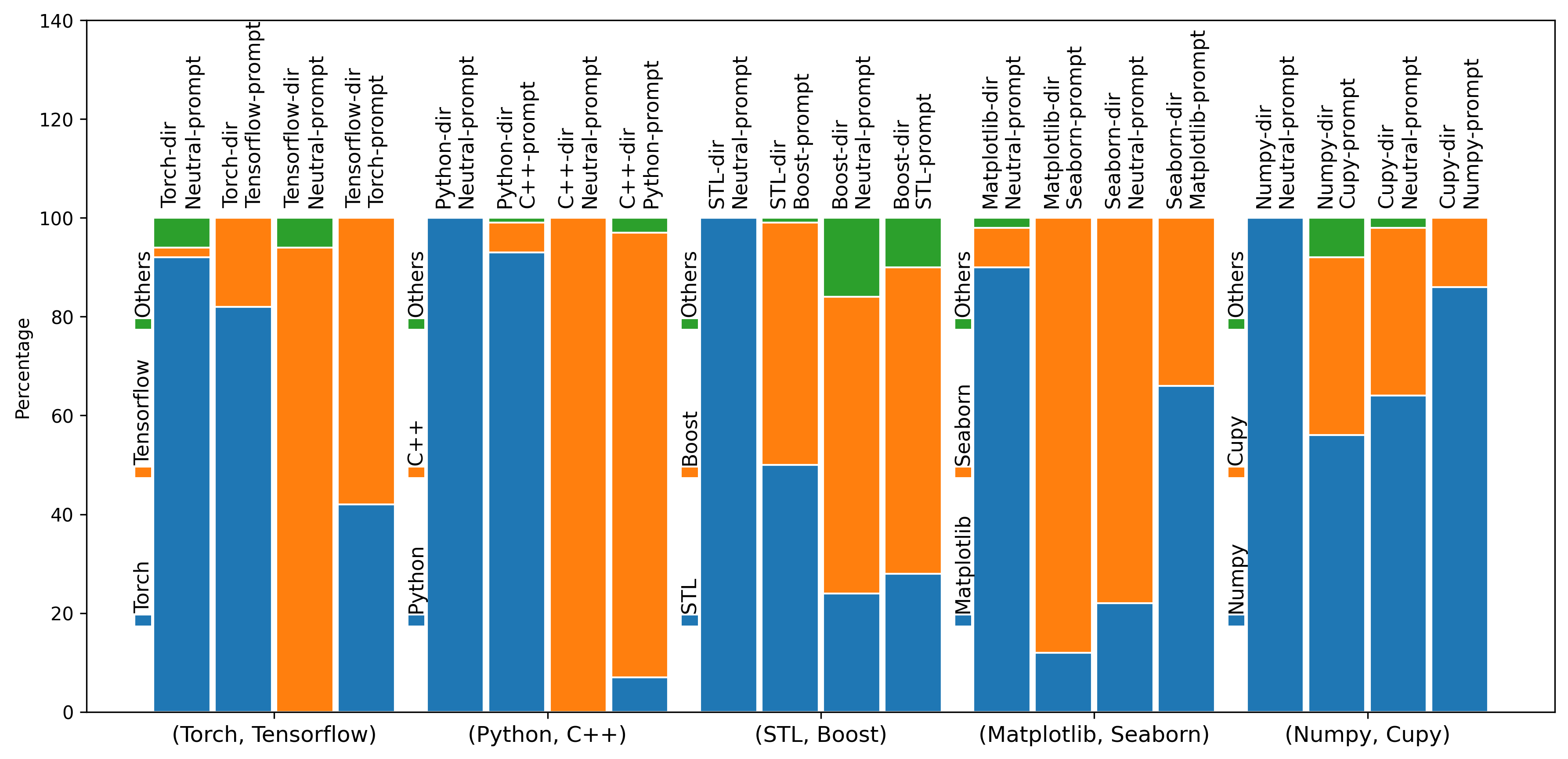}
    \caption{Llama3.1-8B: Output distribution when directional-vector interventions are applied against prompts that request the opposite language or library.}
    \label{fig:llama_conflict}
    \vspace{-12pt}
\end{figure}

We next consider a stricter setting where the prompt explicitly requests the opposite language or library while steering targets the other side, testing whether activation steering can override conflicting instructions.Figures~\ref{fig:codegemma_conflict}, \ref{fig:qwen_conflict}, and \ref{fig:llama_conflict} report results for CodeGemma-7B, Qwen2.5-Coder-7B, and Llama3.1-8B, respectively.

Results show that learned directions remain influential even under prompt conflict. Steering is most effective for common ecosystems, including Torch, TensorFlow, Python, STL, and NumPy, but is weaker and less consistent for rarer alternatives such as Boost, Seaborn, and CuPy. Among the three models, Qwen2.5-Coder-7B exhibits the strongest but most asymmetric override behavior, CodeGemma-7B shows strong effects on several major pairs, and Llama3.1-8B displays the same qualitative trend with more moderate control. Overall, a single activation direction can often compete with or override explicit prompt instructions, though its effectiveness depends on both the model and the target concept.

%% file: sections/conclusion.tex
\section{Conclusion}
We show that programming-language and library preferences in code LLMs can often be steered by adding simple layer-wise activation directions. Across three models and five language/library pairs, these interventions substantially shift generation under neutral prompts and can sometimes override conflicting prompt instructions. At the same time, steering effectiveness varies by model and target ecosystem, and stronger interventions can degrade output quality. Overall, the results suggest that code-style preferences are partly encoded as compact, intervention-friendly structure in activation space, making activation steering a lightweight alternative to prompting or fine-tuning for controlling generation.

%% file: sections/limitations.tex
\section{Limitations}
Our study has several limitations. We evaluate only three open-weight code LLMs and five language/library pairs, so the results may not generalize to other models, ecosystems, or closed-source systems. Most prompts are synthetic, which may not fully reflect real user requests. Our steering method is intentionally simple—a difference-in-means vector applied at a single selected layer—so it may miss more distributed or nonlinear representations. We also use an LLM judge to label target alignment, which does not fully measure code correctness or quality. Finally, stronger steering can degrade output quality, and activation steering may be dual-use beyond the benign code-control setting studied here.

%% file: sections/Acknowledgement.tex
\section{Acknowledgmen}
This work was performed under the auspices of the U.S. Department of Energy (DOE) by Lawrence Livermore National
Laboratory under Contract DE-AC52-07NA27344 (LLNL-CONF-2017025-DRAFT). This material is based upon work
supported by the DOE Office of Science, Advanced Scientific Computing Research program through solicitation DE-FOA0003264, “Advancements in Artificial Intelligence for Science.”

%% file: custom.bib
@article{twist2025study,
  title={A study of llms’ preferences for libraries and programming languages},
  author={Twist, Lukas and Zhang, Jie M and Harman, Mark and Syme, Don and Noppen, Joost and Yannakoudakis, Helen and Nauck, Detlef},
  journal={arXiv preprint arXiv:2503.17181},
  year={2025}
}

@inproceedings{cavs,
  title={Interpretability beyond feature attribution: Quantitative testing with concept activation vectors (tcav)},
  author={Kim, Been and Wattenberg, Martin and Gilmer, Justin and Cai, Carrie and Wexler, James and Viegas, Fernanda and others},
  booktitle={International conference on machine learning},
  pages={2668--2677},
  year={2018},
  organization={PMLR}
}

@article{
    across_layers,
    title={Explaining Explainability: Recommendations for Effective Use of Concept Activation Vectors},
    author={Angus Nicolson and Lisa Schut and Alison Noble and Yarin Gal},
    journal={Transactions on Machine Learning Research},
    issn={2835-8856},
    year={2025},
    url={https://openreview.net/forum?id=7CUluLpLxV},
    note={}
}

@article{steering,
  title={Steering language models with activation engineering},
  author={Turner, Alexander Matt and Thiergart, Lisa and Leech, Gavin and Udell, David and Vazquez, Juan J and Mini, Ulisse and MacDiarmid, Monte},
  journal={arXiv preprint arXiv:2308.10248},
  year={2023}
}

@article{refusal,
  title={Refusal in language models is mediated by a single direction},
  author={Arditi, Andy and Obeso, Oscar and Syed, Aaquib and Paleka, Daniel and Panickssery, Nina and Gurnee, Wes and Nanda, Neel},
  journal={Advances in Neural Information Processing Systems},
  volume={37},
  pages={136037--136083},
  year={2024}
}

@inproceedings{rimsky,
    title = "Steering Llama 2 via Contrastive Activation Addition",
    author = "Rimsky, Nina  and
      Gabrieli, Nick  and
      Schulz, Julian  and
      Tong, Meg  and
      Hubinger, Evan  and
      Turner, Alexander",
    editor = "Ku, Lun-Wei  and
      Martins, Andre  and
      Srikumar, Vivek",
    booktitle = "Proceedings of the 62nd Annual Meeting of the Association for Computational Linguistics (Volume 1: Long Papers)",
    month = aug,
    year = "2024",
    address = "Bangkok, Thailand",
    publisher = "Association for Computational Linguistics",
    url = "https://aclanthology.org/2024.acl-long.828/",
    doi = "10.18653/v1/2024.acl-long.828",
    pages = "15504--15522",
}

@article{stolfo2024improving,
  title={Improving instruction-following in language models through activation steering},
  author={Stolfo, Alessandro and Balachandran, Vidhisha and Yousefi, Safoora and Horvitz, Eric and Nushi, Besmira},
  journal={arXiv preprint arXiv:2410.12877},
  year={2024}
}

@article{lee2024programming,
  title={Programming refusal with conditional activation steering},
  author={Lee, Bruce W and Padhi, Inkit and Ramamurthy, Karthikeyan Natesan and Miehling, Erik and Dognin, Pierre and Nagireddy, Manish and Dhurandhar, Amit},
  journal={arXiv preprint arXiv:2409.05907},
  year={2024}
}

@article{zhao2026odesteer,
  title={ODESteer: A Unified ODE-Based Steering Framework for LLM Alignment},
  author={Zhao, Hongjue and Sun, Haosen and Kong, Jiangtao and Li, Xiaochang and Wang, Qineng and Jiang, Liwei and Zhu, Qi and Abdelzaher, Tarek and Choi, Yejin and Li, Manling and others},
  journal={arXiv preprint arXiv:2602.17560},
  year={2026}
}
